\documentclass[sigconf,natbib=true]{acmart}
\acmConference[Preprint]{}{}
\usepackage{amssymb}
\usepackage{amsmath}
\usepackage{xcolor}
\usepackage{graphicx}
\usepackage{multirow}
\usepackage{subfig}
\usepackage{tabularx}
\usepackage{subcaption}
\newcommand{\BENCHMARK}{REMEDI}
\newcommand{\BENCHMARKFULL}{%
\textbf{R}etention and Unlearning \textbf{E}valuation in \textbf{M}ulti-lab\textbf{e}l Clinical \textbf{D}isease \textbf{I}nference%
}

\begin{document}

\title[REMEDI Benchmark]{REMEDI: A Benchmark for Retention and Unlearning Evaluation in Multi-label Clinical Disease Inference}

\author{Anurag Sharma}
\affiliation{%
  \institution{IIT Kharagpur}
  \country{India}
}

\author{Sai Teja Chunchu}
\affiliation{%
  \institution{IIT Kharagpur}
  \country{India}
}

\author{Prasenjit Mitra}
\affiliation{%
  \institution{Carnegie Mellon University}
  \country{Rwanda}
}

\author{Sandipan Sikdar}
\affiliation{%
  \institution{L3S Research Center,\\ Leibniz University Hannover}
  \country{Germany}
}

\author{Koustav Rudra}
\affiliation{%
  \institution{IIT Kharagpur}
  \country{India}
}
\renewcommand{\shortauthors}{Sharma et al.}

\begin{abstract}
Language models trained for clinical disease inference are trained on patient data, which may include sensitive and private information, and data owners may request the removal of their data from a trained model due to privacy or copyright concerns. However, exactly \emph{unlearning} patient-specific data is intractable, and retraining with minor data removal is resource-intensive. While there exists several machine unlearning methods that can be used, their utility is generally restricted to non-medical domains. Moreover, the existing benchmarks for evaluating such unlearning methods primarily utilize  synthetically curated datasets, which are not truly representative of real-world systems. Hence, the effectiveness of these unlearning methods in the medical domain is largely unclear. To this end, we introduce \textbf{\BENCHMARK}, an extensive benchmark for machine unlearning tailored to {\it multi-label} and {\it multiclass} clinical disease inference, where label correlations, longitudinal structure, and safety constraints make unlearning particularly challenging. 
Unlike the existing benchmarks, \BENCHMARK~conisiders: (1) a relevant  application domain (medical), (2) comprehensive unlearning setups involving diverse sets of forget instances, (3) challenging unlearning scenarios including multi-label and multi-class classification tasks, and (4) evaluation metrics involving performance both in terms of utility and extent of unlearning achieved. \BENCHMARK~is developed using the MIMIC-III clinical database that contains comprehensive clinical data of  patients. Experiments with existing unlearning methods indicate that there exists a trade-off between utility and unlearning performance. They are also largely unsuited to multi-label classification tasks. To facilitate reproducibility, we make our benchmark publicly available\footnote{https://github.com/anuragiiser/REMIDI}. 
\end{abstract}





\maketitle

\section{Introduction}
\label{sec:intro}
In recent years, language models (LMs) have progressed rapidly in the areas of medicine and healthcare  \cite{hassan2024optimizing,nazi2024large}. Training these LMs often involves using vast amounts of text data, which may inadvertently contain sensitive and private information \cite{carlini2021extracting,jonnagaddala2025privacy,ong2024ethical}. In real-world deployment, individuals may exercise their right to have their data removed from a trained language model due to privacy or copyright concerns
or simply because they do not want their data to be used to build models
for other purposes. This ``right to be forgotten'' is a key component of data protection laws \cite{oag2021ccpa} like Europe's General Data Protection Regulation (GDPR) \cite{GDPR}. While retraining the model from scratch after excluding the requested data represents a straightforward solution, this approach is resource-intensive for deep neural networks and foundation models.

Machine unlearning (MU) refers to the selective elimination of specific training samples' influence from a learned model's parameters \cite{cao2015towards,sekhari2021remember,gupta2021adaptive}. We start with a model trained on the complete training dataset, which can be partitioned into two subsets: the \textit{retain set}, which contains data the model should preserve, and the \textit{forget set}, which contains data that should be removed. The goal is to ensure the model performs as if the forget set was never included in its training process \cite{ginart2019making,liu2020federated}. Owing to the infeasibility of re-training from scratch, efficient unlearning algorithms \cite{instance_wise,scrub,gradient_ascent}, have been developed with the objective of producing an unlearned model that approximates the behaviour of a fully retrained model as closely as possible, while requiring only a fraction of the computational cost. However, evaluation of their performance is often limited to synthetically curated datasets and unrealistic unlearning settings, which do not appropriately reflect their effectiveness~\cite{wichert-sikdar-2024-rethinking}. \\
 
In this paper, we introduce a structured and comprehensive benchmark called \textbf{\BENCHMARK} (\BENCHMARKFULL)
to rigorously evaluate machine unlearning methods in clinical multi-label classification settings. The benchmark is built using the MIMIC-III clinical database \cite{johnson2016mimic}. We specifically choose the medical domain, with the unlearning requests involving the removal of patient records. We argue that this is a more realistic unlearning setting, where the effectiveness of an unlearning method can be appropriately measured. 

The task is a multi-label and multi-class classification of electronic health records (EHRs) to International Classification of Diseases (ICD) codes. ICD is developed by the World Health Organization to serve as the global standard for systematic categorization of diseases. Specifically, we use the ICD-9-CM (International Classification of Diseases, Ninth Revision, Clinical Modification) codes \cite{united1991icd}. For instance, \texttt{401} designates `essential hypertension' which is further refined into subcategories such as \texttt{401.0} (malignant essential hypertension), \texttt{401.1} (benign essential hypertension), and so on.

An unlearning benchmark should test unlearning methods across diverse data-deletion scenarios that address real-world needs. Individual patients may request their data be deleted under privacy laws, while an organization might ask for data deletion of entire patient cohorts when data sharing agreements expire or institutional policies change. Moreover, clinical data commonly contains patients with similar characteristics such as demographics, medical histories, or disease presentations. When one patient from such a group requests deletion while others remain in the retain set, the unlearned model must be robust in this particular scenario.
\BENCHMARK~incorporates three unlearning severity levels on the forget set: (1) Distinct Instances, for individual data deletion requests, (2) Concurrent Instances, for realistic scenarios of forget-retain overlap, and (3) Large-scale Instances, for bulk data removal requests, further described in Section \ref{sec:data_splits}. 

After unlearning, the model should behave identically to a model retrained on the retained set, preserve its test performance, and show no trace of the unlearned data. The evaluation framework of \BENCHMARK~operates along two critical dimensions: (i) {\bf Utility}, measured through model performance on forget, retain, and test sets, and (ii) {\bf Privacy}, assessed via membership inference attacks including loss \cite{yeom2018privacy,choi2023towards} to infer whether a each instance of forget set is a member of the model’s training data (refer to Section \ref{subsec:eval_metrics}). 

We evaluate four machine unlearning methods across all three severity levels and compare their performance in terms of model utility and forgetting efficacy. Each baseline represents different algorithmic approach to unlearning: (a) Gradient Ascent \cite{gradient_ascent} employs negative gradients to maximize loss on forget data and directly reverses the learning process, (b) Adversarial Unlearning \cite{instance_wise} utilizes adversarial instance generation and weight importance to selectively diminish the influence of forget samples, (c) Bad Teacher \cite{bad_teacher} uses knowledge distillation with a strategic weak teacher model that has poor performance on forget data, and (d) SCRUB \cite{scrub} fine-tunes the model to maximize disagreement between representations of forget and retain samples. These methods are deliberately selected to evaluate unlearning in the clinical disease inference task. 

\textbf{Our contributions:} ~\BENCHMARK~addresses critical gaps in existing unlearning benchmarks through three key contributions. (1) It 
enables the evaluation of unlearning methods on real-world patient data providing a realistic medical domain testbed. (2) It introduces three unlearning severity levels that capture diverse removal scenarios: distinct instances, concurrent instances, and large-scale instances. (3) The multi-label classification task with correlated disease labels presents significantly greater complexity than existing single-label benchmarks. Multi-label clinical data involves patients with multiple co-occurring disease conditions, which poses unique challenges for unlearning methods. It is a complex task to generate adversarial examples when instances have multiple labels that may be correlated (e.g., Diabetes mellitus (250) and Coronary artery disease (414) often co-occur). Also, negative gradients maximize loss on forget data and disrupt the model's understanding of label correlations that remain valid in the retain set. To selectively remove the influence of specific label combinations without affecting related but distinct disease patterns requires careful disentanglement of learned representations. These challenges make multi-label unlearning a more demanding testbed that reflects the complexity of real-world clinical decision support systems.

Our evaluation framework comprehensively assesses both utility preservation and privacy guarantees through membership inference attacks.

\vspace{-2mm}
\section{Related Work}
Machine unlearning \cite{cao2015towards, bourtoule2021machine} seeks to selectively remove the influence of specific training data from learned models. The majority of existing work has focused on unlearning in supervised classification tasks, where the objective is to eliminate particular training samples while preserving model performance on the remaining data \cite{gradient_ascent,yao2024machine,golatkar2020forgetting}. As machine unlearning methods continue to emerge, the demand for standardized unlearning datasets and rigorous benchmarks in the medical domain have become increasingly urgent. The RWKU benchmark is proposed to unlearn real‐world knowledge from LMs, where the goal is erasing knowledge about real famous people (200 targets) from a model under the setting where neither the “forget corpus” nor the “retain corpus” is fully accessible \cite{rwku}. WHP (Who's Harry Potter) involves fine-tuning the model on a forgetting corpus consisting of the Harry Potter book series \cite{eldan2310s}. The objective of WHP is to render the unlearned model unable to generate content related to Harry Potter,
effectively removing fictional knowledge from the model's parameters. Benchmarks such as TOFU (Task of Fictitious Unlearning) \cite{maini2024tofu} and WMDP (Weapons of Mass Destruction Proxy) \cite{wmdp} adopt this paradigm by providing curated unlearning datasets 
along with carefully designed forget and retain query sets to evaluate unlearning efficacy. TOFU focuses on removing fictitious author-book associations, while WMDP targets the removal of hazardous knowledge related to biosecurity and chemical weapons. MUSE \cite{4muse} provides a comprehensive evaluation framework across six dimensions, having forget quality, model utility, and runtime efficiency. BLUR \cite{blur} evaluates the unlearning of real-world news events and factual knowledge and emphasizes robustness under forget–retain 
overlap and benign relearning, showing that entanglement can mask residual memorization and inflate utility. 

To address the lack of standardized evaluation across different tasks and modalities, MU-Bench \cite{cheng2024mu} was introduced as a multitask
and multimodal benchmark. It is designed to unify the evaluation of unlearning methods by providing standardized deleted sample sets and trained models for a wide range of tasks, including previously unexplored areas like speech and video classification, as well as biomedical relationship extraction. 

Other specialized benchmarks include MMDU-Bench \cite{zhangmmdu} for multi-modal knowledge graph unlearning and datasets like MUFAC and MUCAC for multi-class classification \cite{choi2023towards}. However, most evaluation frameworks and benchmarks have been developed for non-medical domains. There is a significant gap in understanding how unlearning methods perform under the unique constraints and complexities of clinical data with multi-label classifications. 

Table~\ref{tab:benchmark_comparison} presents a qualitative comparison of existing unlearning benchmarks with \BENCHMARK.

While existing benchmarks have advanced machine unlearning research, they exhibit critical limitations that \BENCHMARK~addresses. Benchmarks like TOFU, WMDP, and RWKU use synthetic datasets in non-medical domains and focus on concept-level unlearning (removing entire topics), whereas \BENCHMARK~leverages real-world patient EHR data from MIMIC-III and focuses on patient-level removal, where individual records must be selectively eliminated without degrading diagnostic capability on clinically similar cases. In
addition, \BENCHMARK~uniquely combines multi-label ICD-9 disease classification with three clinically realistic severity levels, including forget-retain overlap scenarios where patients to be forgotten share characteristics with retained patients, and rigorous privacy verification via membership inference attacks, positioning it as the first comprehensive benchmark tailored to clinical machine unlearning requirements.
\begin{table}[htb]
\setlength{\tabcolsep}{2.5pt}
\renewcommand{\arraystretch}{1}
\centering
\scriptsize
% \footnotesize
\label{tab:benchmark_comparison}
\begin{tabular}{l|c|c|c|c|c|c|c|c}
\hline
\textbf{Feature} & \textbf{MUSE} & \textbf{WMDP} & \textbf{TOFU} & \textbf{WHP} & \textbf{RWKU} & \textbf{BLUR} & \textbf{MU} & \textbf{\BENCHMARK} \\
\hline
\hline
\multicolumn{9}{c}{\textit{Domain Characteristics}} \\
\hline
Medical domain & & & & & & & & \checkmark \\
\hline
Real-world patient data & & & & & & & & \checkmark \\
\hline
\hline
\multicolumn{9}{c}{\textit{Task Complexity}} \\
\hline
Multi-label classification & & & & & & & \checkmark & \checkmark \\
\hline
\hline
\multicolumn{9}{c}{\textit{Unlearning Granularity}} \\
\hline
Instance-level & \checkmark & & \checkmark & & & \checkmark & \checkmark & \checkmark \\
\hline
%Class-level & \checkmark & \checkmark & & & \checkmark & \checkmark & \checkmark & \\
%\hline
\hline
\multicolumn{9}{c}{\textit{Evaluation Metrics}} \\
\hline
Forget quality & \checkmark & \checkmark & \checkmark & \checkmark & \checkmark & \checkmark & \checkmark & \checkmark \\
\hline
Retain utility & \checkmark & \checkmark & \checkmark & \checkmark & \checkmark & \checkmark & \checkmark & \checkmark \\
\hline
Privacy (MIA) & \checkmark & & \checkmark & & & \checkmark & & \checkmark \\
\hline
\hline
\multicolumn{9}{c}{\textit{Benchmark Features}} \\
\hline
Multiple severity levels & & & & & & & & \checkmark \\
\hline
Forget-retain overlap & & & & & & \checkmark & & \checkmark\\
\hline
Open-source dataset & \checkmark & \checkmark & \checkmark & \checkmark & \checkmark & \checkmark & \checkmark & \checkmark \\
\hline
\end{tabular}
\vspace{1em}
\caption{Comparison with existing machine unlearning benchmarks across key features. Compare following benchmarks: MUSE \cite{4muse}; WMDP \cite{wmdp}; TOFU \cite{maini2024tofu}; WHP \cite{eldan2310s}; RWKU \cite{rwku}; BLUR \cite{blur}; MU \cite{zhangmmdu}; \BENCHMARK{} = Ours.}
\label{tab:benchmark_comparison}
\vspace{-7mm}
\end{table}
\section{The \BENCHMARK~Benchmark}
We elaborate on each component in the following subsections.
\subsection{Task Definition}
\label{subsec:task_definition}

We use electronic health record (EHR) data, from MIMIC-III clinical database \cite{johnson2016mimic} which includes structured clinical notes and patient histories, for disease inference.

The task involves mapping clinical notes to the most precise ICD codes. Given an EHR, the model predicts associated diseases represented as ICD-9-CM (International Classification of Diseases, Ninth Revision, Clinical Modification) codes \cite{united1991icd}. The main objective of clinical outcome prediction is to support medical professionals in differential diagnosis, the systematic process of distinguishing between multiple conditions that share similar presentations \cite{van-aken-etal-2021-clinical}. Our dataset contains ten ICD-9-CM code classes. We formulate the task as a multi-label classification problem where each patient admission may be associated with multiple concurrent diagnoses. Unlike symptom-based inference tasks, the disease codes in our benchmark are directly extracted from clinical documentation rather than inferred from symptom descriptions alone. Formally, given an input EHR $x_i$, the model $f_\theta$ produces a prediction $\hat{y}_i \in \{0,1\}^K$, where $K=10$ represents the number of ICD-9-CM code classes, and each element $\hat{y}_i^{(k)}$ indicates the presence or absence of the $k$-th disease.

\subsection{Task Setting} \label{subsection:task_setting}

\paragraph{Definition.}
Let \( \mathcal{D} = \{(x_i, y_i)\}_{i=1}^n \) be the original training set, where $x_i$ is the EHR of a patient and $y_i$ is a set ICD-9 disease codes i.e., the true labels. 

Let \( \mathcal{D}_f \subset \mathcal{D} \) be the forget set, and let \( \mathcal{D}_r = \mathcal{D} \setminus \mathcal{D}_f \) be the retain set. 
 
\begin{figure}[tb]
    \centering
\includegraphics[width=0.8\columnwidth]{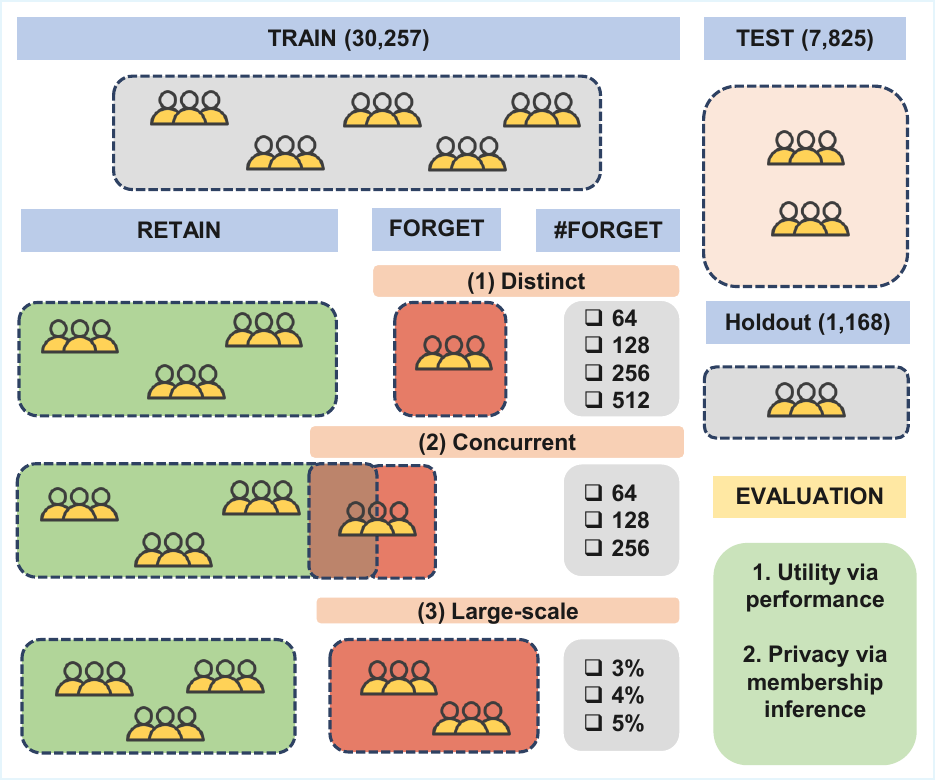}
    \caption{\BENCHMARK~provides three kinds of forget sets to measure robustness to varying sizes of forget sets: (1) Distinct instances, (2) Concurrent instances, and (3) Large-scale instances. \#forget represents the number of instances in the forget set. Large-scale instances are \% of the train set.}
    \label{fig:forget_splits}
\end{figure}

Traditional unlearning tasks \cite{wmdp,maini2024tofu}, typically provide a forget set $\mathcal{D}_f \in \mathcal{D}$, which constitutes a subset of the training data $\mathcal{D}$ that contains samples to be removed. An unlearning method aims to update the model to approximate its behavior as if it had been trained exclusively on the $\mathcal{D}_r$. Formally, given an unlearning objective, a model $f_\theta$ with parameters $\theta$ is updated via an unlearning algorithm, to yield an unlearned model with modified parameters $\theta'$. In the medical domain, the forget set $\mathcal{D}_f$ comprises sensitive patient records protected under privacy regulations such as GDPR \cite{GDPR} and HIPAA \cite{gostin2009beyond}. We propose a \emph{task-invariant instance-level unlearning}, where the unlearning request targets instances rather than classes, and the original prediction task and label space are preserved, and no class is entirely removed. Figure~\ref{fig:forget_splits} provides an overview of the approach. The training algorithm $\mathcal{A}$ gives the original model \( f_{\theta} = \mathcal{A}(\mathcal{D}) \). An unlearning algorithm $\bar{\mathcal{A}}$ produces an unlearned model \(f_{\theta'} = \bar{\mathcal{A}}(\mathcal{D}_f, f_{\theta})\)
 such that \( f_{\theta'} \) behaves as closely as the retrained model \( f_{\tilde{\theta}} = \mathcal{A}(\mathcal{D}_r) \).
 %\sandipan{What about MIA?}\done~ 
 Additionally, the unlearned model must satisfy privacy constraints. Formally, let \( \mathcal{D}_h \) denote a holdout set disjoint from \( \mathcal{D} \). Perfect unlearning requires that the loss distribution of \( f_{\theta'} \) on \( \mathcal{D}_f \) is indistinguishable from its distribution on \( \mathcal{D}_h \):
\[
\mathcal{L}(f_{\theta'}(x_f), y_f) \stackrel{d}{\approx} \mathcal{L}(f_{\theta'}(x_h), y_h), \quad \forall (x_f, y_f) \in \mathcal{D}_f, (x_h, y_h) \in \mathcal{D}_h
\]
This property is verified through membership inference attack (MIA), where an adversary trained to distinguish between these distributions should achieve accuracy close to random guessing (0.5), indicating that forget samples leave no detectable trace in the model.

\subsection{Data Source and Construction}
A robust medical unlearning benchmark must apply to diverse biomedical language models, which involve the data that represents knowledge widely present across such models. To this end, we utilize the MIMIC-III and MIMIC-IV (Medical Information Mart for Intensive Care III) Clinical Database \cite{johnson2016mimic}, a freely available repository comprising de-identified health-related data from over 40,000 patients who received care in critical care units at Beth Israel Deaconess Medical Center between 2001 and 2012. The database contains comprehensive clinical information, including patient demographics, bedside vital sign measurements, laboratory test results, clinical procedures, medication records, caregiver notes, diagnostic imaging reports, and mortality outcomes. For our benchmark construction, we focus on discharge summaries and their associated diagnostic outcomes. Discharge summaries provide rich, unstructured clinical narratives that synthesize a patient's hospital course, making them ideal for evaluating language model performance on clinical reasoning tasks. We consider all diagnoses associated with each hospital admission, which are encoded as ICD-9 codes in the MIMIC-III database. Following \cite{pmlr-v68-choi17a}, we group ICD-9 diagnosis codes from their original 4-digit granularity into broader 3-digit categories to reduce task complexity while maintaining clinical meaningfulness. But we still have 1266 categories after this. 

We follow the preprocessing steps outlined by \citet{van-aken-etal-2021-clinical} to ensure that our task reflects information available at the time of patient admission. We filter discharge summaries to retain only those sections that are typically documented upon admission, which include: \textit{Chief Complaint, (History of) Present Illness, Medical History, Admission Medications, Allergies, Physical Exam, Family History, and Social History}. To construct a manageable yet representative multi-label classification task, we identify the ten most frequently occurring disease categories in the dataset. The categories are: Hypertension (401), Cardiac arrhythmia (427), Heart failure (428), Electrolyte imbalance (276), Diabetes mellitus (250), Coronary artery disease (414), Hyperlipidemia (272), Anemia (285), Respiratory failure (518), Acute kidney failure (584). These ten diseases represent the most common conditions encountered in critical care settings and provide sufficient sample diversity for robust evaluation. Each patient admission is then labeled with a binary vector indicating the presence or absence of each of these disease codes, resulting in a multi-label dataset where patients may have multiple diagnoses. 

It is important to note that the class distribution in MIMIC-III is highly skewed, with some diseases appearing far more frequently than others. Including all 1266 categories would make the prediction task intractable due to severe class imbalance and sparse labels. By focusing on the ten most frequent disease categories, we balance task difficulty with clinical relevance while ensuring sufficient sample representation for robust unlearning evaluation.

% \vspace{-2cm}

\subsection{Forget Data}
\label{sec:data_splits}
The benchmark comprises 39,250 EHRs with labeled ICD-9-CM disease codes in total, divided into four parts: (1) training set (30,257 EHRs), (2) validation set (1,168 EHRs), (3) test set (6,325 EHRs), and (4) holdout set (1,500 EHRs). A clear illustration of these data splits is presented in Figure~\ref{fig:forget_splits}. The holdout set is reserved for privacy evaluation and is never used during model training or unlearning procedures. To comprehensively evaluate unlearning methods across realistic clinical scenarios, we define three severity levels of forget sets, each reflecting different real-world data removal requests:

\textbf{(1) Distinct Instances (\(\mathcal{D}_{f\_distinct} \), Low Severity):} This setting simulates individual patient data removal requests, where forget instances are randomly sampled from the training set with no overlap with the retain set. We construct four forget set sizes: 64, 128, 256, and 512 instances. To ensure robustness and statistical reliability, we provide three independent folds for each size for cross-validation of unlearning performance. This scenario reflects cases where individual patients withdraw consent or request data deletion under GDPR provisions.

\textbf{(2) Concurrent Instances (\(\mathcal{D}_{f\_concurrent} \), Medium Severity):} This configuration evaluates unlearning under challenging conditions where forget and retain sets contain overlapping information. Here, half of the instances in each forget set share similar clinical characteristics (e.g., demographic profiles or temporal admission patterns) with instances in the retain set. We create three forget set sizes: 64, 128, and 256 instances, where 32, 64, and 128 instances, respectively, overlap with the retain set. This design tests the unlearning algorithm's ability to selectively remove specific patient data while preserving the model's diagnostic capability on clinically similar cases, which is a critical requirement when patients with shared conditions request data deletion.

\textbf{(3) Large-Scale Instances (\(\mathcal{D}_{f\_large} \), High Severity):} This setting addresses scenarios where organizational or governmental mandates require bulk data removal, such as regulatory compliance actions and institutional data breaches. We construct forget sets containing 3\%, 4\%, and 5\% of the training data (907, 1,210, and 1,513 instances, respectively). This severity level stress-tests the unlearning efficiency and utility preservation capabilities of an unlearning algorithm when faced with substantial data removal requests.

Figure \ref{fig:forget_splits} provides a thorough visualization of the benchmark structure, illustrating the relationships between training, forget, retain, validation, test, and holdout sets across all severity levels.

\subsection{Evaluation Metrics}
\label{subsec:eval_metrics}

As described in Section \ref{subsection:task_setting}, an unlearned model should behave as if it had never been exposed to the forget set, exhibiting behavior similar to a model retrained exclusively on the retain set. Formally, we desire $f_{\theta'} \approx f_{\tilde{\theta}}$. Our evaluation considers two critical properties: (i) \textit{Utility via performance} on forget, retain, and test sets, and (ii) \textit{Privacy via membership inference}.

\noindent \textbf{(1) Utility via Performance}
An ideal unlearning method must preserve the model's generalization capability on the original task and eliminate the influence of the forget set. In multi-label classification, the macro-F1 score is a commonly used metric that treats all labels with equal importance that makes it particularly suitable for medical diagnosis tasks where each disease category is clinically significant. We compute the macro-F1 score across three evaluation sets: test, retain, and forget. $F_{\text{test}}$ evaluates generalization performance on test data. $F_{\text{retain}}$ checks if the unlearned model performance preservation of the retain set. $F_{\text{forget}}$ evaluates the degradation of performance on the forget set.

Formally, for a set of \(K\) disease labels, the macro-F1 score is computed as:
\begin{equation}
F_{\text{macro}} = \frac{1}{K} \sum_{k=1}^{K} \frac{2 \cdot \text{Precision}^{(k)} \cdot \text{Recall}^{(k)}}{\text{Precision}^{(k)} + \text{Recall}^{(k)}}
\end{equation}
where \(\text{Precision}^{(k)}\) and \(\text{Recall}^{(k)}\) are the precision and recall for the \(k\)-th label, respectively.

An ideal unlearned model should have same $F_{\text{test}}$, $F_{\text{retain}}$, $F_{\text{forget}}$ scores to the retrained model.\\

\noindent \textbf{(2) Privacy via Membership Inference Attack:} To verify that patient records have been effectively unlearned, we assess whether the model retains identifiable knowledge of the forget set. It is essential that the unlearned model does not leak membership information indicating that $\mathcal{D}_f$ was part of $\mathcal{D}_{\text{train}}$. We utilize, membership inference attacks (MIA) \cite{miacarlini2022membership} to quantify this privacy leakage. The key intuition behind MIA is that models exhibit lower loss values on training samples compared to unseen data. For the originally trained model $f_\theta$, the loss $\mathcal{L}(f_\theta(x_{\text{train}}), y_{\text{train}})$ for samples from $\mathcal{D}_{\text{train}}$ is typically lower than the loss $\mathcal{L}(f_\theta(x_{\text{holdout}}), y_{\text{holdout}})$ for samples from the holdout set $\mathcal{D}_{\text{holdout}}$. We follow \cite{choi2023towards,4muse} to quantify the privacy metric. We train a binary logistic regression classifier $\Theta(\cdot)$ to distinguish between loss values of forget set samples $x_f \in \mathcal{D}_f$ and holdout set samples $x_h \in \mathcal{D}_{\text{holdout}}$. The classifier attempts to predict membership based on the loss values produced by the unlearned model:
\begin{equation}
\Theta(x) = 
\begin{cases}
1 & \text{if } x \in \mathcal{D}_f \\
0 & \text{if } x \in \mathcal{D}_{\text{holdout}}
\end{cases}
\end{equation}

Perfect unlearning corresponds to a classification accuracy of 0.5, which indicates that the forget set samples are statistically indistinguishable from holdout samples. We define the Privacy via membership inference as:
\begin{equation}
\text{Privacy via membership inference} = 2 \cdot \left| \text{accuracy}(\Theta) - 0.5 \right|
\end{equation}
The scaling factor of 2 ensures that the value is between 0 and 1, which makes it easier to interpret. A lower MIA score indicates stronger privacy guarantees, with a score of 0 representing perfect unlearning, where the forget set samples are completely indistinguishable from unseen data.

\begin{table}[t]
\centering
\setlength{\tabcolsep}{2.5pt}
\renewcommand{\arraystretch}{1}
\begin{tabular}{|c|cccc|cccc|}
\hline
\multirow{3}{*}{Methods} & \multicolumn{4}{c}{BioLinkBERT}                                                                                 & \multicolumn{4}{|c|}{BioBERT}                                                                                     \\ \cline{2-9} 
                         & \multicolumn{3}{c|}{Utility (F1) ($\uparrow$)}                                                     & \multirow{2}{*}{\shortstack{MIA\\($\downarrow$)}} & \multicolumn{3}{c|}{Utility (F1) ($\uparrow$)}                                                     & \multirow{2}{*}{\shortstack{MIA\\($\downarrow$)}} \\ \cline{2-4} \cline{6-8}
                         & \multicolumn{1}{c|}{Test} & \multicolumn{1}{c|}{Forget} & \multicolumn{1}{c|}{Retain} &                          & \multicolumn{1}{c|}{Test} & \multicolumn{1}{c|}{Forget} & \multicolumn{1}{c|}{Retain} &                          \\ \hline
Original                 & \multicolumn{1}{c|}{0.60} & \multicolumn{1}{c|}{0.63}   & \multicolumn{1}{c|}{0.63}   & 0.03                     & \multicolumn{1}{c|}{0.54} & \multicolumn{1}{c|}{0.57}   & \multicolumn{1}{c|}{0.57}   & 0.01                     \\ \hline
Retrain                  & \multicolumn{1}{c|}{0.63} & \multicolumn{1}{c|}{0.62}   & \multicolumn{1}{c|}{0.61}   & 0.07                     & \multicolumn{1}{c|}{0.58} & \multicolumn{1}{c|}{0.55}   & \multicolumn{1}{c|}{0.58}   & 0.02                     \\ \hline
GA+R                       & \multicolumn{1}{c|}{0.07} & \multicolumn{1}{c|}{0.07}   & \multicolumn{1}{c|}{0.08}   & 0.11                     & \multicolumn{1}{c|}{0.06} & \multicolumn{1}{c|}{0.06}   & \multicolumn{1}{c|}{0.06}   & 0.08                     \\ \hline
GA+                      & \multicolumn{1}{c|}{0.63} & \multicolumn{1}{c|}{0.59}   & \multicolumn{1}{c|}{0.66}   & 0.01                     & \multicolumn{1}{c|}{0.57} & \multicolumn{1}{c|}{0.51}   & \multicolumn{1}{c|}{0.60}   & 0.03                     \\ \hline
AU                       & \multicolumn{1}{c|}{0.52} & \multicolumn{1}{c|}{0.49}   & \multicolumn{1}{c|}{0.55}   & 0.07                     & \multicolumn{1}{c|}{0.28} & \multicolumn{1}{c|}{0.29}   & \multicolumn{1}{c|}{0.32}   & 0.05                     \\ \hline
AU+WR                    & \multicolumn{1}{c|}{0.53} & \multicolumn{1}{c|}{0.51}   & \multicolumn{1}{c|}{0.55}   & 0.01                     & \multicolumn{1}{c|}{0.32} & \multicolumn{1}{c|}{0.31}   & \multicolumn{1}{c|}{0.36}   & 0.03                     \\ \hline
BT                       & \multicolumn{1}{c|}{0.47} & \multicolumn{1}{c|}{0.47}   & \multicolumn{1}{c|}{0.49}   & 0.12                     & \multicolumn{1}{c|}{0.45} & \multicolumn{1}{c|}{0.46}   & \multicolumn{1}{c|}{0.47}   & 0.11                     \\ \hline
SCRUB                    & \multicolumn{1}{c|}{0.62} & \multicolumn{1}{c|}{0.67}   & \multicolumn{1}{c|}{0.65}   & 0.02                     & \multicolumn{1}{c|}{0.56} & \multicolumn{1}{c|}{0.58}   & \multicolumn{1}{c|}{0.59}   & 0.10                     \\ \hline
\end{tabular}
\caption{Evaluation results before and after unlearning 64 instances of \(\mathcal{D}_{f\_distinct} \) the forget set using BioLinkBERT and BioBERT models. }
\label{tab:distinct}
\vspace{-7mm}
\end{table}

\begin{table}[t]
\centering
\setlength{\tabcolsep}{2.5pt}
\renewcommand{\arraystretch}{1}
\begin{tabular}{|c|cccc|cccc|}
\hline
\multirow{3}{*}{Methods} & \multicolumn{4}{c|}{BioLinkBERT}                                                                                 & \multicolumn{4}{c|}{BioBERT}                                                                                     \\ \cline{2-9} 
                         & \multicolumn{3}{c|}{Utility (F1) ($\uparrow$)}                                                     & \multirow{2}{*}{\shortstack{MIA\\($\downarrow$)}} & \multicolumn{3}{c|}{Utility (F1) ($\uparrow$)}                                                     & \multirow{2}{*}{\shortstack{MIA\\($\downarrow$)}} \\ \cline{2-4} \cline{6-8}
                         & \multicolumn{1}{c|}{Test} & \multicolumn{1}{c|}{Forget} & \multicolumn{1}{c|}{Retain} &                          & \multicolumn{1}{c|}{Test} & \multicolumn{1}{c|}{Forget} & \multicolumn{1}{c|}{Retain} &                          \\ \hline
Original                 & \multicolumn{1}{c|}{0.60} & \multicolumn{1}{c|}{0.64}   & \multicolumn{1}{c|}{0.63}   & 0.04                     & \multicolumn{1}{c|}{0.54} & \multicolumn{1}{c|}{0.58}   & \multicolumn{1}{c|}{0.57}   & 0.06                     \\ \hline
Retrain                  & \multicolumn{1}{c|}{0.58} & \multicolumn{1}{c|}{0.60}   & \multicolumn{1}{c|}{0.61}   & 0.02                     & \multicolumn{1}{c|}{0.57} & \multicolumn{1}{c|}{0.57}   & \multicolumn{1}{c|}{0.58}   & 0.06                     \\ \hline
GA                       & \multicolumn{1}{c|}{0.06} & \multicolumn{1}{c|}{0.06}   & \multicolumn{1}{c|}{0.06}   & 0.04                     & \multicolumn{1}{c|}{0.05} & \multicolumn{1}{c|}{0.04}   & \multicolumn{1}{c|}{0.05}   & 0.04                     \\ \hline
GA+R                      & \multicolumn{1}{c|}{0.64} & \multicolumn{1}{c|}{0.60}   & \multicolumn{1}{c|}{0.67}   & 0.05                     & \multicolumn{1}{c|}{0.60} & \multicolumn{1}{c|}{0.56}   & \multicolumn{1}{c|}{0.64}   & 0.04                     \\ \hline
AU                       & \multicolumn{1}{c|}{0.20} & \multicolumn{1}{c|}{0.20}   & \multicolumn{1}{c|}{0.24}   & 0.03                     & \multicolumn{1}{c|}{0.07} & \multicolumn{1}{c|}{0.07}   & \multicolumn{1}{c|}{0.07}   & 0.04                     \\ \hline
AU+WR                    & \multicolumn{1}{c|}{0.40} & \multicolumn{1}{c|}{0.40}   & \multicolumn{1}{c|}{0.44}   & 0.06                     & \multicolumn{1}{c|}{0.08} & \multicolumn{1}{c|}{0.09}   & \multicolumn{1}{c|}{0.09}   & 0.06                     \\ \hline
BT                       & \multicolumn{1}{c|}{0.48} & \multicolumn{1}{c|}{0.47}   & \multicolumn{1}{c|}{0.50}   & 0.02                     & \multicolumn{1}{c|}{0.42} & \multicolumn{1}{c|}{0.41}   & \multicolumn{1}{c|}{0.45}   & 0.03                     \\ \hline
SCRUB                    & \multicolumn{1}{c|}{0.62} & \multicolumn{1}{c|}{0.65}   & \multicolumn{1}{c|}{0.65}   & 0.01                     & \multicolumn{1}{c|}{0.58} & \multicolumn{1}{c|}{0.61}   & \multicolumn{1}{c|}{0.60}   & 0.05                     \\ \hline
\end{tabular}
\caption{Evaluation results before and after unlearning 128 instances of \(\mathcal{D}_{f\_concurrent} \) the forget set using BioLinkBERT and BioBERT models.}
\label{tab:concurrent}
\vspace{-7mm}
\end{table}
\section{Experimental Setup}
\begin{figure*}[tb]
    \centering
    \includegraphics[width=0.75\linewidth]{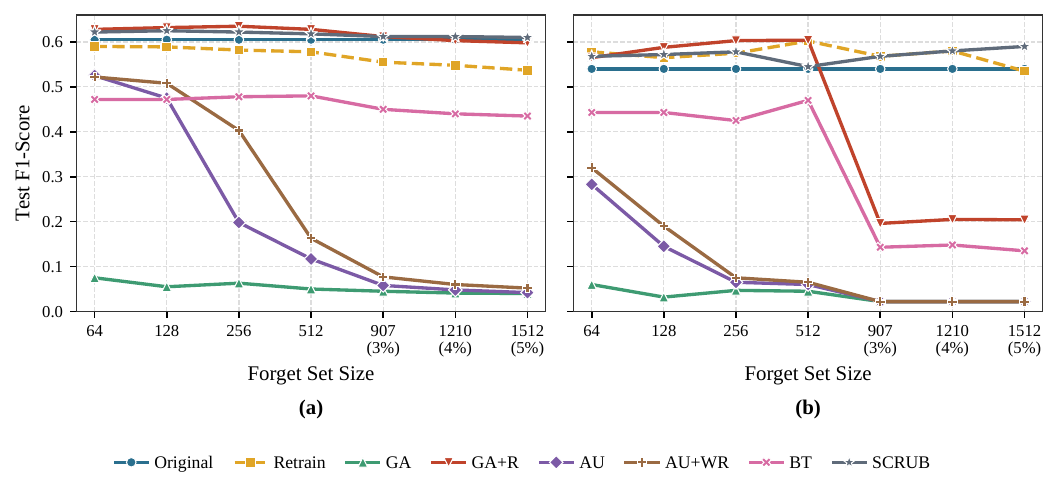}
    \caption{Utility evaluation across forget set sizes for distinct and large-scale unlearning scenarios. (a) BioBERT: Only SCRUB retains performance across all forget set sizes in large-scale scenarios. (b) BioLinkBERT: GA+R, BT, and SCRUB maintain stable utility across varying $\mathcal{D}_{\text{forget}}$ sizes.}
    \label{fig:lineplot}
    \vspace{-5mm}
\end{figure*}
\subsection{Models}
\label{subsec:models}
We conduct unlearning experiments on two biomedical language models: BioLinkBERT \cite{yasunaga2022linkbert} and BioBERT \cite{lee2020biobert}, both pretrained on large-scale biomedical corpora and well-suited for clinical text understanding. To obtain the original trained model $f_\theta$, we fine-tune each base model on the task mentioned in Section \ref{subsec:task_definition}. We utilize the training set $\mathcal{D}_{\text{train}}$ with the validation set $\mathcal{D}_{\text{val}}$ for hyperparameter selection and early stopping. After fine-tuning, we evaluate machine unlearning algorithms that take as input the trained model $f_\theta$, the forget set $\mathcal{D}_f$, and the retain set $\mathcal{D}_r$, to produce an unlearned model $f_{\theta'}$ that approximates a model retrained on $\mathcal{D}_r$. In particular, unlearning methods do not access the test set $\mathcal{D}_{\text{test}}$ or holdout set $\mathcal{D}_{\text{holdout}}$ during the unlearning phase.

\subsection{Baseline Unlearning Methods}
\label{subsec:baselines}
We evaluate four representative unlearning methods from different algorithmic families, each adapted to the multi-label classification setting. To ensure training stability we employ the AdamW optimizer with gradient clipping and a safe weighted binary cross-entropy loss function. For each unlearning strategy, we train for multiple epochs and select the model checkpoint that achieves the lowest macro-F1 score on the validation set. The hyperparameter details are present in Github repo. The baselines are as follows:

(a) {\bf Gradient Ascent (GA):} GA \cite{gradient_ascent} forgets by maximizing the loss on the forget set. We use (i) {\bf GA} (forget-only): $\min_{\theta}-\mathcal{L}_f(\theta)$, where $\mathcal{L}_f$ is the BCE loss on forget samples; and (ii) {\bf GA+R} (retain-aware): $\mathcal{L}=(1-\lambda)\mathcal{L}_r-\lambda\mathcal{L}_f$, with $\lambda\in[0,1]$ trading off retention and forgetting.

{\bf Adversarial Unlearning (AU):} AU \cite{instance_wise} performs instance-wise forgetting via adversarial perturbations of forget samples. We consider (i) {\bf AU}: $\mathcal{L}=\lambda_r\mathcal{L}_{\text{adv}}-\lambda_f\mathcal{L}_{\text{forget}}$, and (ii) {\bf AU+WR}: $\mathcal{L}=\lambda_r\mathcal{L}_{\text{adv}}-\lambda_f\mathcal{L}_{\text{forget}}+\lambda_w\mathcal{L}_{\text{reg}}$, where $\mathcal{L}_{\text{reg}}$ regularizes important weights.

{\bf Bad Teacher (BT):} BT \cite{bad_teacher} uses distillation to forget selectively: a student mimics a weak teacher on forget samples and a strong teacher on retain samples. We optimize\\
$\mathcal{L}_{\text{BT}}=\mathrm{KL}\!\left(p_s \,\|\, y\,p_u+(1-y)\,p_f\right)$,
where $y\in\{0,1\}$ indicates forget membership and temperature scaling stabilizes training.

{\bf SCRUB:} SCRUB \cite{scrub} alternates between increasing divergence on forget samples and preserving behavior on retain samples. The objective is
$\mathcal{L}=\gamma \mathcal{L}_{\text{cls}}+\alpha \mathcal{L}_{\text{div}}$,
where $\mathcal{L}_{\text{div}}$ is also distillation-based divergence from the original model.
\section{Results}
\label{sec:results}

We compare four unlearning methods and their variants against the original model and the retrained model across utility and privacy metrics. The retrained model is fine-tuned exclusively on the retain set $\mathcal{D}_r$ that serves as the gold standard, representing the ideal behavior of a perfectly unlearned model. Tables \ref{tab:distinct} and \ref{tab:concurrent} present the main experimental results for BioLinkBERT and BioBERT across distinct (\(\mathcal{D}_{f\_distinct} \)) and concurrent (\(\mathcal{D}_{f\_concurrent} \)) forget set configurations, respectively. Our evaluation assesses whether unlearning methods can approximate retrained model performance while maintaining computational efficiency and strong privacy guarantees. We have shown additional results on the different splits forget sets in the repository.

\textbf{Overall Performance Analysis.}
GA ascent leads to severe performance degradation, effectively nullifying the model’s prediction capabilities. When we incorporate a retain set (GA+R), which mitigates the performance degradation, it preserves the utility comparable to full retraining. AU demonstrates moderate effectiveness. Standard adversarial training destabilizes learned representations, whereas the weight-regularized variant alleviates this effect by emphasizing parameter importance, thereby preserving essential diagnostic knowledge while enhancing privacy retention. BT achieves moderate utility preservation but suffers in case of large-scale forget sets. It indicates that residual teacher influence constrains the model’s ability to fully eliminate forget-set information. Overall, SCRUB best utility–privacy balance, giving performance near that of retrained models while maintaining strong privacy metric. Across all methods, BioLinkBERT consistently surpasses BioBERT in both utility preservation and privacy maintenance, attributable to its entity-linked pretraining that captures richer clinical semantics. In contrast, BioBERT exhibits higher sensitivity to unlearning interventions, emphasizing the influence of pretraining objectives and architectural design on unlearning effectiveness.

\textbf{Scalability Analysis Across Forget Set Sizes.} To assess the robustness of unlearning methods under varying sizes of data removal requests, we evaluate utility preservation across different severity levels of forget set. We analysis F1 score for the forget set size ranging from 64 instances to 1,512 instances (5\% of training data. Figure \ref{fig:lineplot} illustrates the test F1 scores for all methods as forget set size increases. For BioBERT (Figure \ref{fig:lineplot}a), SCRUB emerges as the only method that consistently maintains performance close to the retrained model across all forget set sizes, including large-scale scenarios. GA causes complete utility collapse regardless of forget set size, while its retain-aware variant (GA+R) shows moderate stability but degrades noticeably beyond 512 instances. Adversarial methods (AU and AU+WR) exhibit early degradation even at smaller forget set sizes, struggling to balance forgetting with utility preservation. Bad Teacher maintains reasonable performance up to medium-scale forgetting but experiences significant drops in large-scale scenarios. BioLinkBERT (Figure \ref{fig:lineplot}b) demonstrates greater overall resilience to unlearning interventions. Three methods: GA+R, BT, and SCRUB maintain stable performance across all forget set sizes, closely tracking the retrained model's performance. GA again fails, while adversarial approaches show progressive utility degradation as forget set size increases. These results highlight a critical finding: most unlearning methods are optimized for small-scale data removal but struggle to maintain utility when faced with large-scale unlearning requests.
\vspace{-2mm}
\section{Conclusion}
\label{sec:conclu}
% \section{Conclusion}
\BENCHMARK{}~frames machine unlearning as a dual objective by preserving diagnostic utility while removing patient-level information. We present a multi-label setting that reveals failure modes of the unlearning methods that perform perfectly on non-medical or synthetic benchmarks. Our results quantify this for distinct instances as GA+R can collapse BioLinkBERT utility. However, methods such as, SCRUB achieve a markedly stronger utility-privacy balance, and the unlearned model remains stable under concurrent forgetting. The scaling study further indicates that apparent success at small deletions does not reliably transfer to larger forget sets, with BioBERT maintaining near-retrained utility across sizes only under SCRUB while other baselines degrade. These findings motivate several open directions: multi-label--aware objectives that respect clinically meaningful label correlations, overlap-robust deletion that separates an individual's contribution from similar retained cohorts, privacy evaluation beyond loss-based membership inference, and audit-efficient pipelines that support repeated deletion requests without full retraining.

\bibliographystyle{ACM-Reference-Format}
\bibliography{reference}

\end{document}